\def\year{2021}\relax
\documentclass[letterpaper]{article} 
\usepackage{aaai21}  
\usepackage{times}  
\usepackage{helvet} 
\usepackage{courier}  
\usepackage[hyphens]{url}  
\usepackage{graphicx} 
\usepackage{natbib}  
\usepackage{caption} 
\usepackage{latexsym}
\usepackage{comment}

\usepackage{multirow} 
\usepackage{amsmath, amsfonts, amsthm, amssymb}
\usepackage{subfigure} 
\usepackage[inline]{enumitem}

\newtheoremstyle{mysubtask}%
{}{}{\normalfont}{}{\scshape}{.\,}{ }{}
\theoremstyle{mysubtask}
\newtheorem{subtask}{Subtask}

\newtheoremstyle{mytask}%
{}{}{\normalfont}{}{\scshape}{.\,}{ }{}
\theoremstyle{mytask}
\newtheorem{task}{Task}

\newtheoremstyle{mydef}%
{}{}{\normalfont}{}{\scshape}{.\,}{ }{}
\theoremstyle{mydef}

\date{}

\title{VisualMRC: Machine Reading Comprehension on Document Images}

\author{
    Ryota Tanaka\thanks{The first two authors have equal contribution.},
    Kyosuke Nishida\footnotemark[1], 
    Sen Yoshida\\
}
\affiliations{
    NTT Media Intelligence Laboratories, NTT Corporation \\
    \{ryouta.tanaka.rg, kyosuke.nishida.rx, sen.yoshida.tu\}@hco.ntt.co.jp
}

\begin{document}

\maketitle

\begin{abstract}
Recent studies on machine reading comprehension have focused on text-level understanding but have not yet reached the level of human understanding of the visual layout and content of real-world documents. In this study, we introduce a new visual machine reading comprehension dataset, named VisualMRC, wherein given a question and a document image, a machine reads and comprehends texts in the image to answer the question in natural language. Compared with existing visual question answering (VQA) datasets that contain texts in images, VisualMRC focuses more on developing natural language understanding and generation abilities. It contains 30,000+ pairs of a question and an abstractive answer for 10,000+ document images sourced from multiple domains of webpages. We also introduce a new model that extends existing sequence-to-sequence models, pre-trained with large-scale text corpora, to take into account the visual layout and content of documents. Experiments with VisualMRC show that this model outperformed the base sequence-to-sequence models and a state-of-the-art VQA model. However, its performance is still below that of humans on most automatic evaluation metrics. The dataset will facilitate research aimed at connecting vision and language understanding.
\end{abstract}

\section{Introduction}

Creating intelligent agents that can answer questions as well as people can is a long-cherished goal of artificial intelligence.
To achieve this goal, machine reading comprehension (MRC), a challenge to enable a machine to read and comprehend natural language texts so that it can answer questions, has received much attention~\cite{RajpurkarZLL16,RajpurkarJL18}. 
The MRC capability can be of value to users if it can be employed by automated assistants such as  customer-service chatbots on e-commerce websites~\cite{CuiHWTDZ17} or assistant systems for reading professional literature~\cite{HongWJZW19}. 
Here, most real-world documents are provided in non-plain text formats (e.g., HTML and PDF). However, current studies in MRC almost exclusively focus on text-level understanding, while neglecting the visual layout and content (text appearance, tables, charts, etc.) of the documents.
Visual question answering (VQA) on images containing a few words~\cite{SinghNSJCBPR19,BitenTMBRJVK19} has recently been studied as a challenging task that lies at the intersection of vision and language understanding. However, these learning tasks do not focus on document understanding. They cannot
be used to develop the ability to make a machine visually read and comprehend real-world documents.

\begin{figure}[t!]
    \centering
    \includegraphics[width=.47\textwidth]{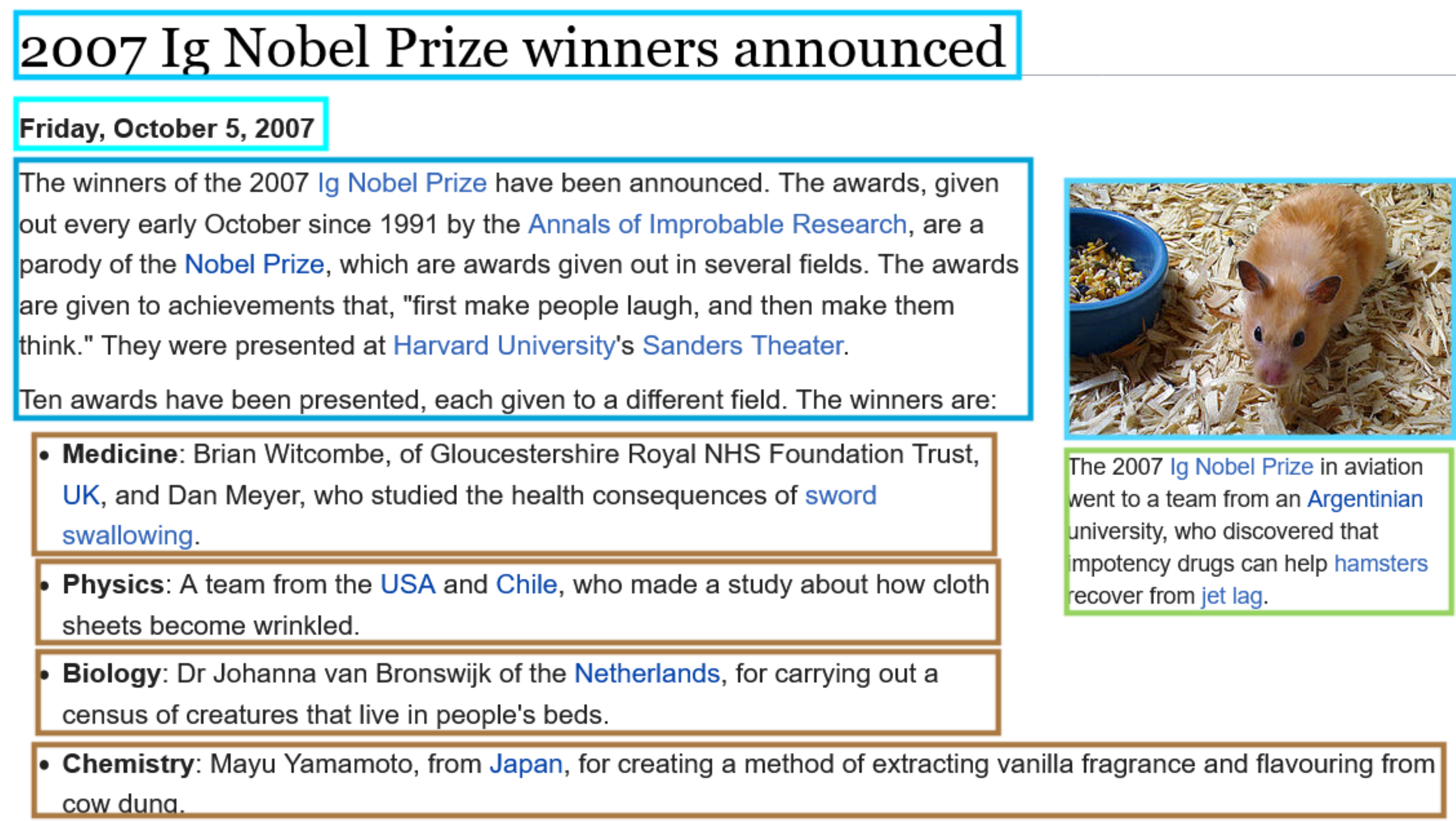}
    \footnotesize
    \tabcolsep=3pt
    \scalebox{1.0}{
    \begin{tabular}{p{23em}}
    \\
        {\bf Q}: Who were the winners of the Ig Nobel prize for Biology and Chemistry? \\
        {\bf A}: The winner of the Ig Nobel prize for biology was Dr Johanna van Bronswijk, and the winner for Chemistry was Mayu Yamamoto. \\ 
         \\ 
    \end{tabular}
    }
    \caption{Example from our VisualMRC dataset.
    The dataset provides regions-of-interest (ROIs) with semantic classes in addition to QA pairs and document images.  
    E.g., a bounding box colored in brown indicates a list. 
    The image was sourced from  \url{https://en.wikinews.org/wiki/2007_Ig_Nobel_Prize_winners_announced}.}
\label{fig:example_dataset}
\end{figure}

To move towards more advanced 
vision and language understanding, we have created a new dataset\footnote{
We will release further information about this dataset at \url{https://github.com/nttmdlab-nlp/VisualMRC}.}, called Visual Machine Reading Comprehension (VisualMRC), wherein given a question and a document image, a machine reads and comprehends texts in an image to  answer the question in natural language. As shown in Figure~\ref{fig:example_dataset}, the task demands a rich set of abilities as varied as understanding the document layout and visual representations of the text and non-text objects and extracting relations between objects, in addition to natural language understanding (NLU) and natural language generation (NLG). 
Compared with DocVQA~\cite{Mathew_2021_WACV}, which is a concurrent work of VQA on document images, our dataset differs in that it provides a number of images sourced from various contemporary webpages and it provides long abstractive answers that can be used for evaluating the NLU and NLG abilities on document images.

Our main contributions are as follows.
\begin{itemize}
\item We introduce a novel visual machine reading comprehension dataset (VisualMRC) containing QA 
pairs that require a machine to read and reason about texts in the document image.
Our dataset is currently the only one dataset that focuses on generative QA on document images.

		

\item We propose a new model that allows for transferring the NLU and NLG abilities of sequence-to-sequence models, pre-trained on text corpora, to the VisualMRC task.

\item Our model outperformed existing state-of-the-art 
VQA model~\cite{HuSDR20} and sequence-to-sequence models that we used as the base models~\cite{RaffelSRLNMZLL20,LewisLGGMLSZ20}
on the VisualMRC dataset.

\end{itemize}

\section{Existing Vision and Language QA Datasets}

\subsubsection{VQA on images containing a few words.}

VQA, in which a document in MRC takes the form of an image, has been studied intensively~\cite{AntolALMBZP15,
GoyalKSBP17
}.  
Recently, a number of VQA datasets with text in images, annotated using optical character recognition (OCR), have been released.
The focus of most \textit{text-based} VQA datasets is to reason and answer questions about text in natural daily scenes.
VizWiz-VQA~\cite{
Gurari0SGLGLB18} 
consists of questions originating from blind people who each took a picture using a mobile phone.
TextVQA~\cite{SinghNSJCBPR19}, ST-VQA~\cite{BitenTMBRJVK19}, and 
EST-VQA~\cite{WangLSNLJCHW20} are crowd-sourced 
datasets on everyday scenes.
Moreover, some datasets focus on different types of images. For instance, OCR-VQA~\cite{MishraSSC19} is a dataset containing images of book covers;
FigureQA~\cite{KahouMAKTB18} and DVQA~\cite{KaflePCK18} 
are datasets containing 
diagrams and charts.
What is most different about our 
dataset in comparison with the ones mentioned above is that its images contain more words. 
Our work is focused more on developing the NLU ability on documents in which multiple pieces of text and visual content are laid out.

\subsubsection{VQA on document images.}

Similarly to VisualMRC, DocVQA~\cite{Mathew_2021_WACV} has proposed a dataset for VQA that requires reading and reasoning about document images. However, there are important differences in design choices, as follows:
\begin{enumerate*}[label=(\roman*)]
\item  VisualMRC contains a number of different images sourced from multiple domains, while the images of  DocVQA are from a single source, the UCSF Industry Documents Library.
\item  VisualMRC consists of contemporary born-digital webpages, while most of the documents in DocVQA are from the 1960--2000 period, containing handwritten or typewritten words.
\item The images of VisualMRC contain a minimum of three natural language sentences, while there is no guarantee that natural language sentences are included in the images of DocVQA. 
\item VisualMRC provides long abstractive answers, while 
DocVQA provides SQuAD-like extractive and short answers from a single span of the text in the document image.
\end{enumerate*}

Natural Questions~\cite{KwiatkowskiPRCP19} is an MRC dataset that provides HTML documents, and we may be able to use it as a VQA dataset by creating document images with HTML rendering; however, even state-of-the-art models like RikiNet~\cite{LiuGFYCJLD20} do not use visual information.

\subsubsection{Multi-modal question answering.}

Multi-modal question answering takes 
both textual and visual information as input contexts, which is different from text-based VQA that takes only an image as the input context.
%
TQA~\cite{KembhaviSSCFH17} is comprised of middle-school science lessons containing diagrams and text.
RecipeQA~\cite{YagciogluEEI18} provides cooking recipes with images and text.
The motivation behind these studies is similar to ours, but 
the visual information in the text such as 
the document layout is dropped from the text in these datasets. 
The focus of our research is to enable machines to handle the same visual input as humans do when they read real-world documents.

Moreover, some of these datasets are in the setting of multiple-choice QA that allows for accurate evaluation. However, in terms of application scenarios, it is highly cost to collect answer candidates to answer open-domain questions. For this reason, we believe that generative settings are important, even if they are difficult to evaluate. 

\section{The VisualMRC Task and Dataset}


We first define the VisualMRC task and then describe the data collection concerning the task's input and output.

\subsection{Task Overview}

We present VisualMRC, a new vision and language task to read and comprehend texts given as a document image.

First, the end-to-end task is simply defined as:
\begin{task}[End-to-end VisualMRC]
\label{prob:prob}
Given a question $q$ and a document image $I$, a model generates an answer $a$. 
\end{task}
\noindent The VisualMRC task is a generative MRC task such as NarrativeQA~\cite{KociskySBDHMG18},
in which the answer is not limited to word spans in the context.
The understanding of the image can be decomposed into two sub-tasks:
\begin{subtask}[Region-Of-Interest (ROI) detection]
\label{prob:roi}
Given an image $I$, a model detects a set of ROIs.
Each ROI $r_i$ consists of a bounding box $b_i$ and a semantic class label $l_i$.
\end{subtask}
\begin{subtask}[OCR]
\label{prob:ocr}
Given a ROI $r_i$, a model detects a set of word objects within the region. Each word object consists of a bounding box $b_{i,j}$ and a form $w_{i,j}$.
\end{subtask}

\subsection{Dataset Collection}

We describe how we collected images, ROIs, OCR words, and question-answer pairs. Our dataset provides \textbf{ground-truth ROIs} annotated by humans 
and OCR words for each ROI as the outputs of \textsc{Subtasks} 1 and 2. It also provides \textbf{relevant ROIs} that are required to answer each question.

\subsubsection{Document image collection.}

First, we collected 5,599 full screenshots of webpages in 35 domains licensed under creative commons from January to March 2020. 
Then, we asked 94 crowdworkers to determine if each page included any content that is usable for creating QA pairs and to annotate the content (as a document image $I$) 
with a bounding box. 
They were allowed to annotate multiple pieces of content in a screenshot but were not allowed to overlap the bounding boxes. Finally, seven crowdworkers validated the annotated content. In total, 10,197 images were collected.
We defined content that is suitable as a document image as follows.
\begin{enumerate*}[label=(\roman*)]
\item No handwritten text is allowed: only machine-printed text.
\item The content is preferred to contain both pictures and texts, but this is not strictly required.
\item The content must contain a minimum of three natural language sentences, preferably no more than 2-3 paragraphs.
\item The content has to contain information at least two of the 
classes described in the next subsection.
\end{enumerate*}

\subsubsection{Ground-truth ROI annotation.}

\begin{figure}[t!]
    \centering
    \includegraphics[width=.4\textwidth]{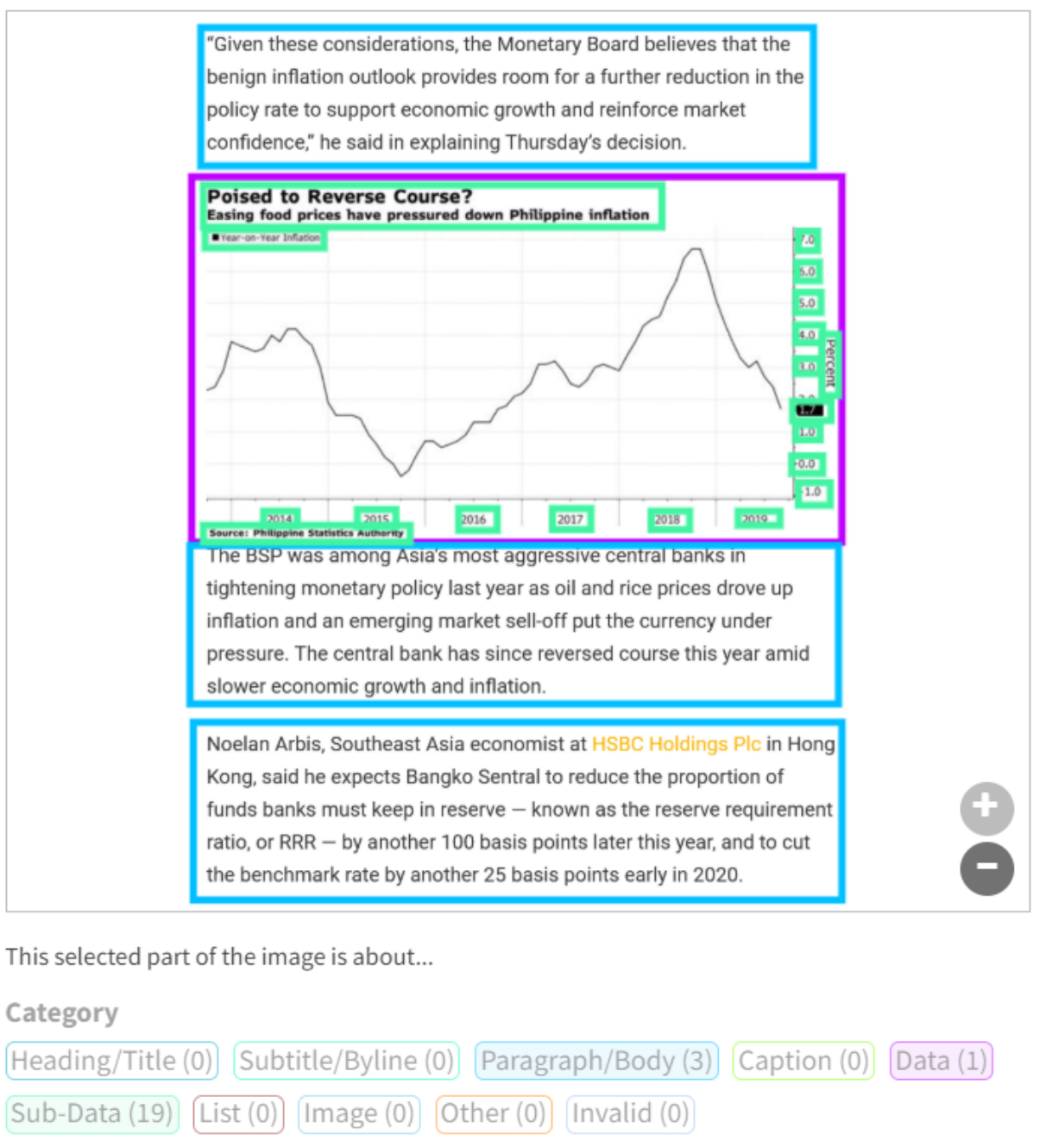}
    \caption{Screenshot of ROI annotation by crowdworkers.}
    \label{fig:annot}
\end{figure}

45 crowdworkers were asked to indicate specific objects (ROI $r_i$ in \textsc{Subtask} 1) in the given image $I$ by annotating bounding boxes $b_i$ around the objects and classifying them into nine classes $l_i$.
Figure~\ref{fig:annot} shows a screenshot showing crowdworkers' ROI annotation by selecting a class among the nine classes for each ROI.


We defined the nine ROI classes as follows. 
\begin{itemize}
\item \textbf{Heading/Title} The title or caption of a page, chapter, etc.
\item \textbf{Subtitle/Byline} 
The secondary or subordinate title of a page or  
a line of text giving the author's name.
\item \textbf{Paragraph/Body} The main 
text that would be read.
\item \textbf{Picture} 
The picture or image that contains no text or data.
\item \textbf{Caption} The text placed next 
an image, data, etc. 
that provides or explains information about an image or data.
\item \textbf{List} Typically bulleted lists, where each bullet is not a full sentence.
\item \textbf{Data} Tables, charts, graphs, infographic, or other figures with data or information.
\item \textbf{Sub-data} The text placed inside of the Data region.
\item \textbf{Other} Any other text that does not fit in the other categories.
\end{itemize}%

\subsubsection{OCR execution.}

We 
extracted words (bounding box $b_{i,j}$ and form $w_{i,j}$ in \textsc{Subtask} 2) from each ROI except the data regions (because we have sub-data regions for texts in a data region)
by using the Tesseract OCR system~\cite{Smith07}.

\subsubsection{QA pair collection.}

495 crowdworkers 
created three unique questions $q$, and their generative answers $a$ for each image $I$, where questions should ask about the written content and there should be only one answer to each question. 

\subsubsection{Relevant ROI annotation.}
The crowdworker that created a question-answer pair also chose the relevant ROIs (required to answer the question) among the ground-truth ROIs. 
79 crowdworkers validated the created data.

\subsubsection{Data split.}

We split the dataset into training, development, and test sets, in terms of URL domain; the datasets contain 21,015, 2,839, and 6,708 questions, respectively.

\begin{figure*}[t!]
    \centering
    \subfigure[Number of questions with a particular question length.]{
        \includegraphics[width=.29\textwidth]{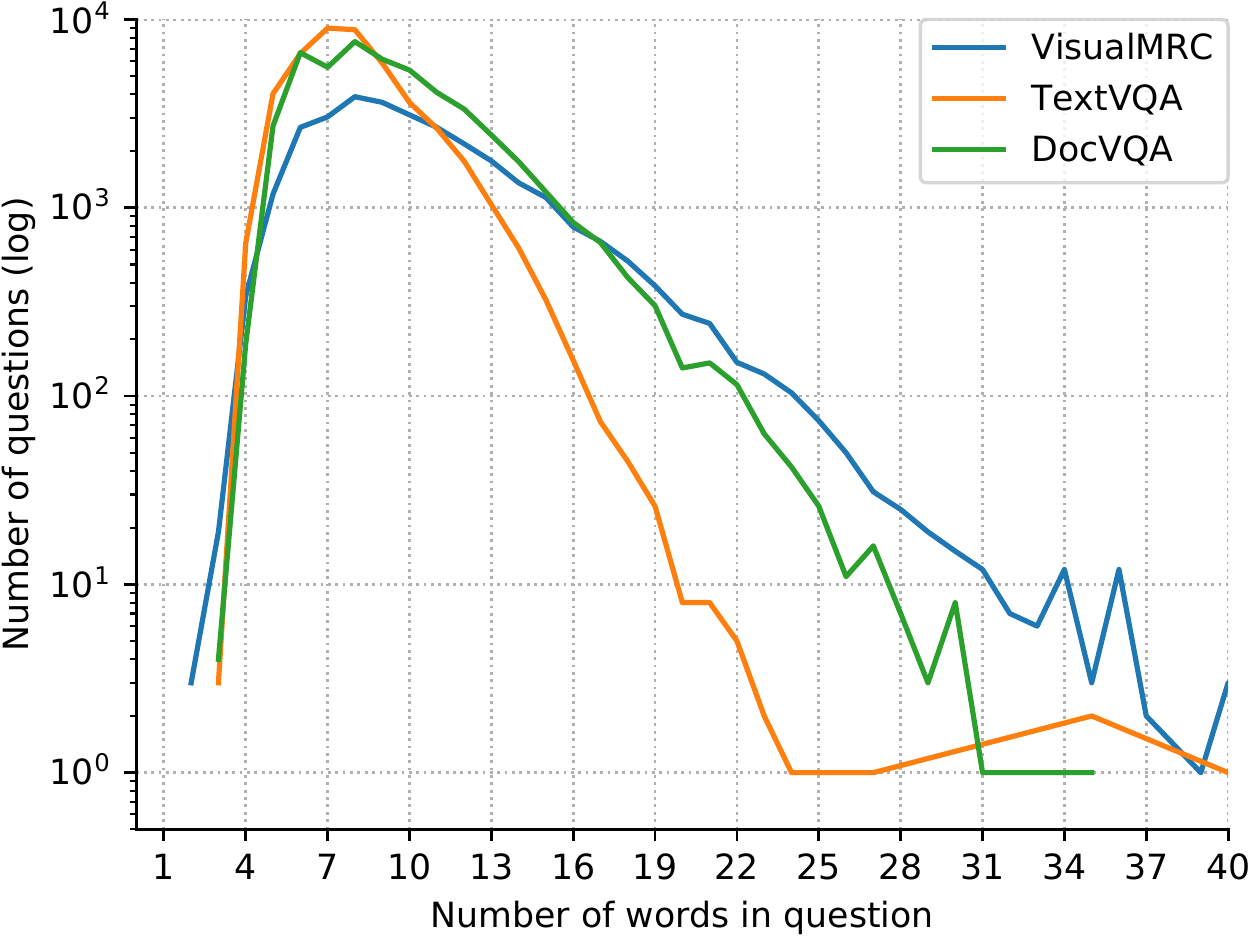}
    }
    \hspace{1em}
    \subfigure[Number of answers with a particular answer length.]{
        \includegraphics[width=.29\textwidth]{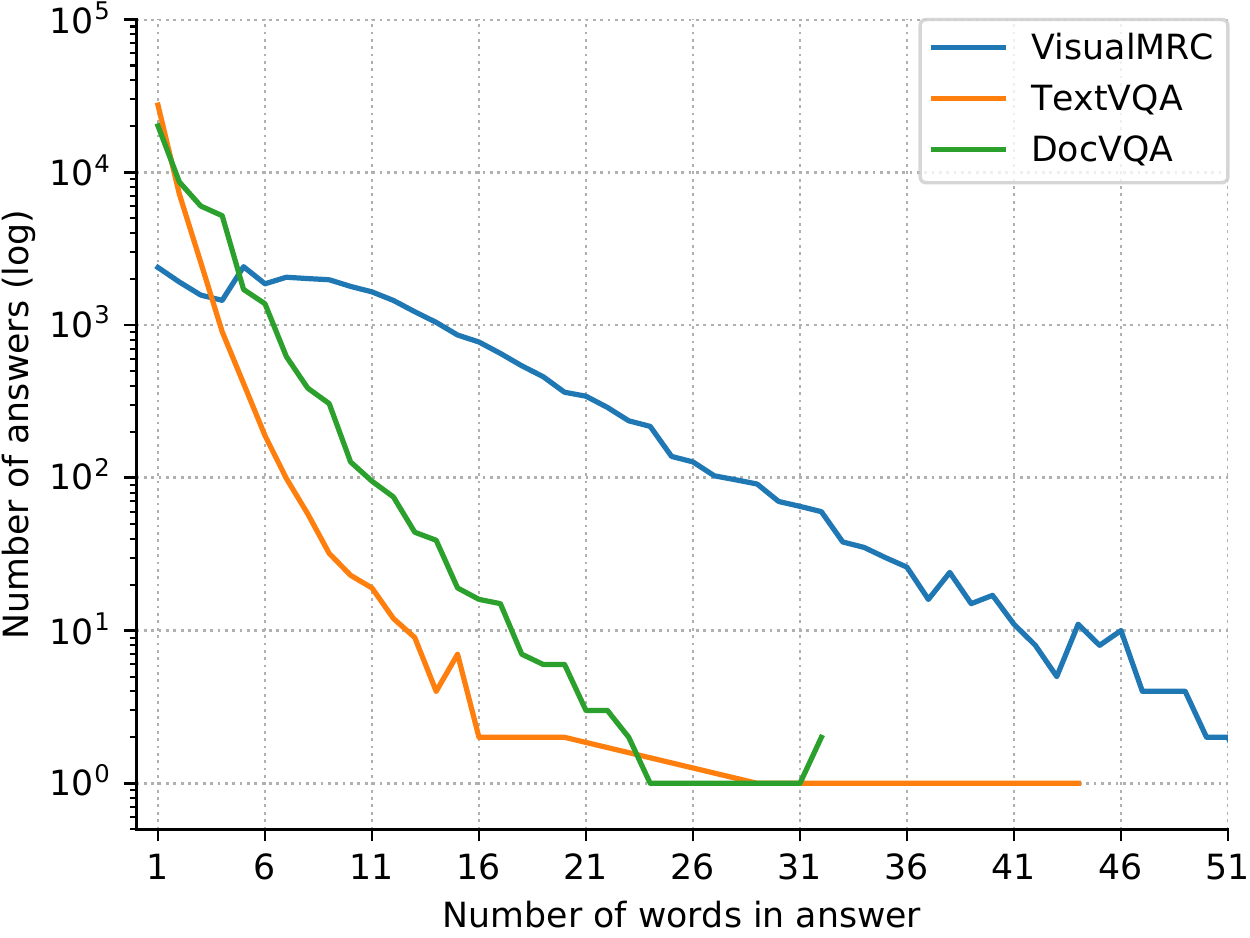}
    }
    \hspace{1em}
    \subfigure[Number of document images with a particular number of OCR words.]{
        \includegraphics[width=.29\textwidth]{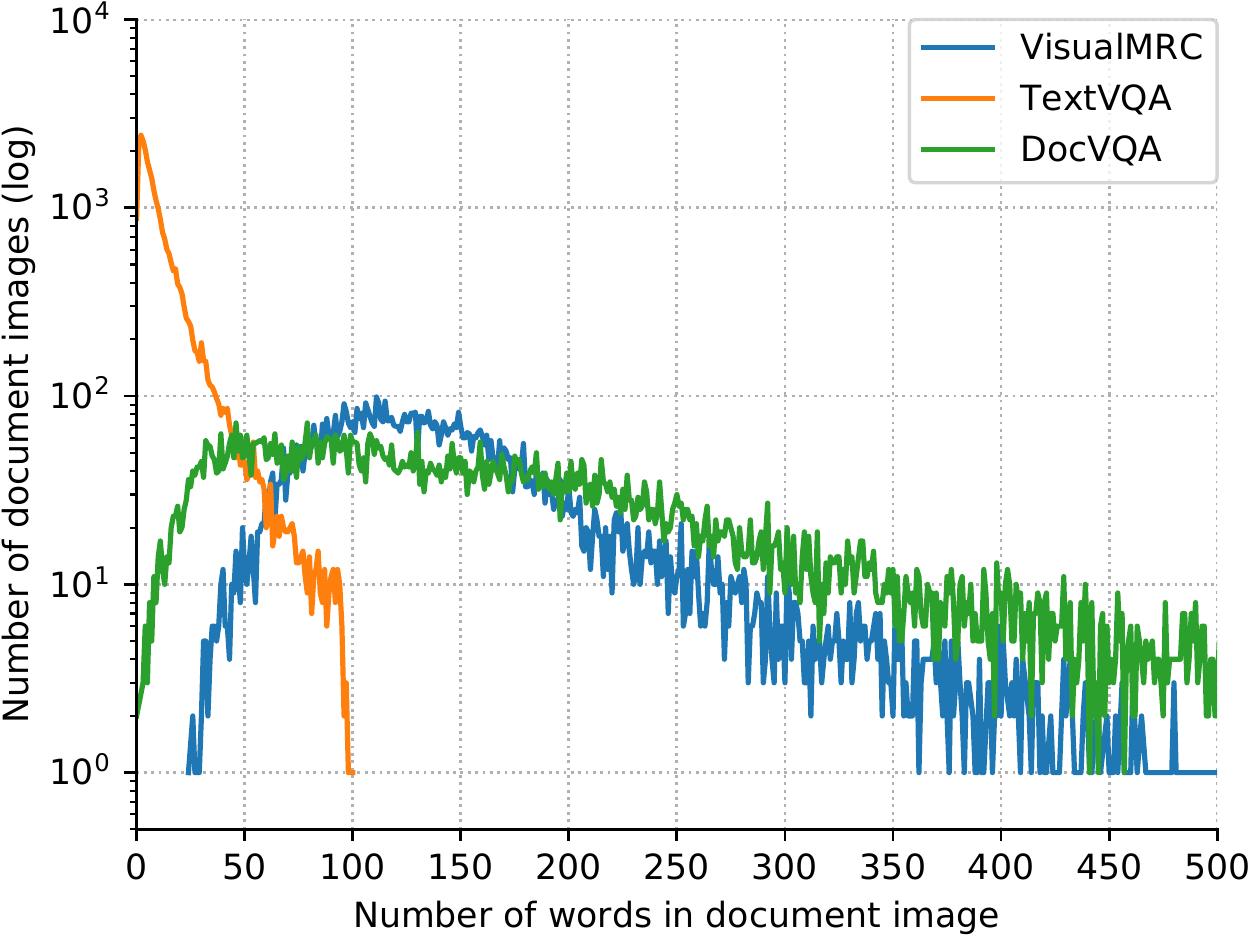} 
    }
    \subfigure[Word cloud for questions.]{
    \includegraphics[width=.29\textwidth]{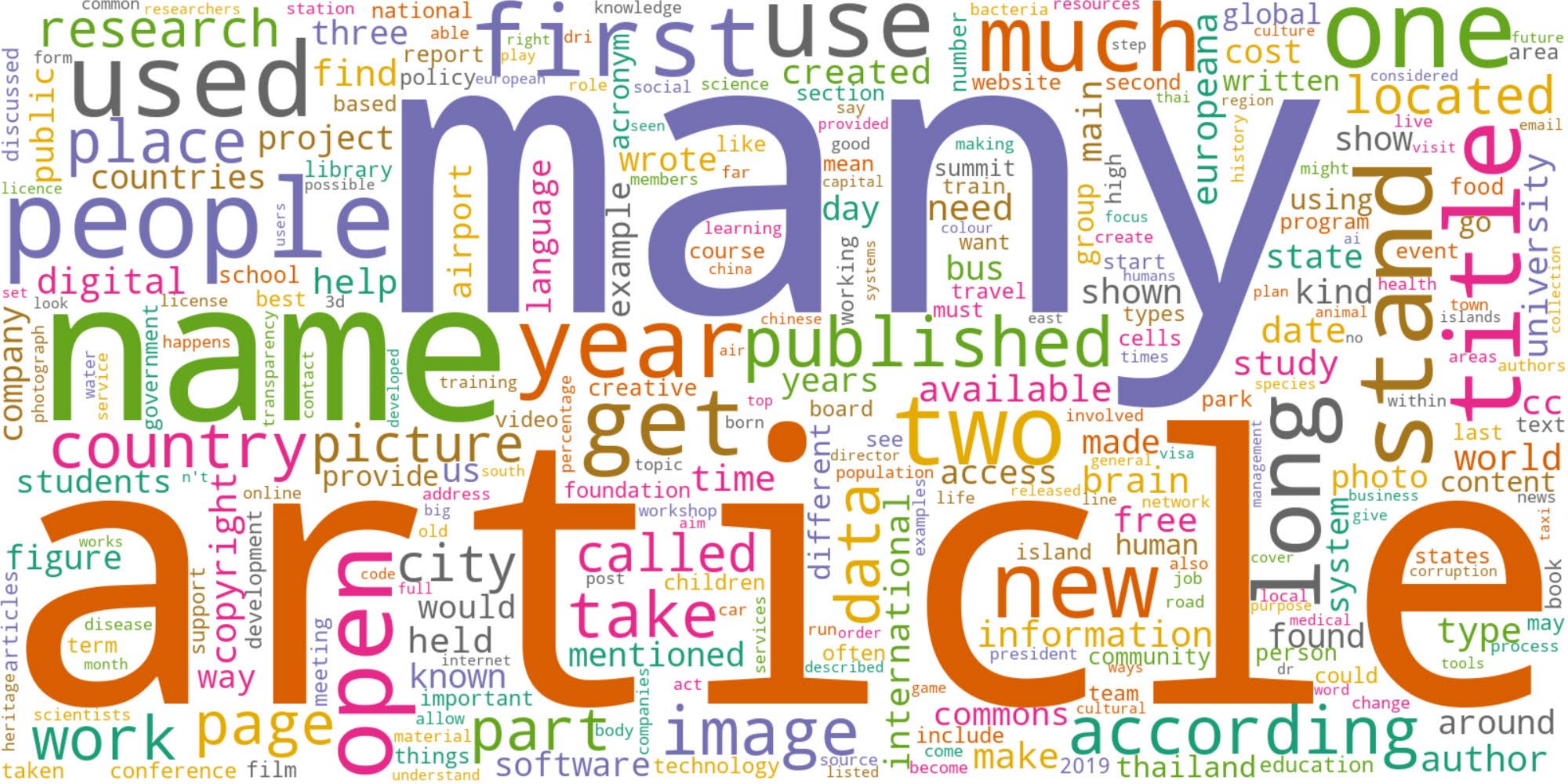}
    }
    \hspace{1em}
    \subfigure[Word cloud for answers.]{
    \includegraphics[width=.29\textwidth]{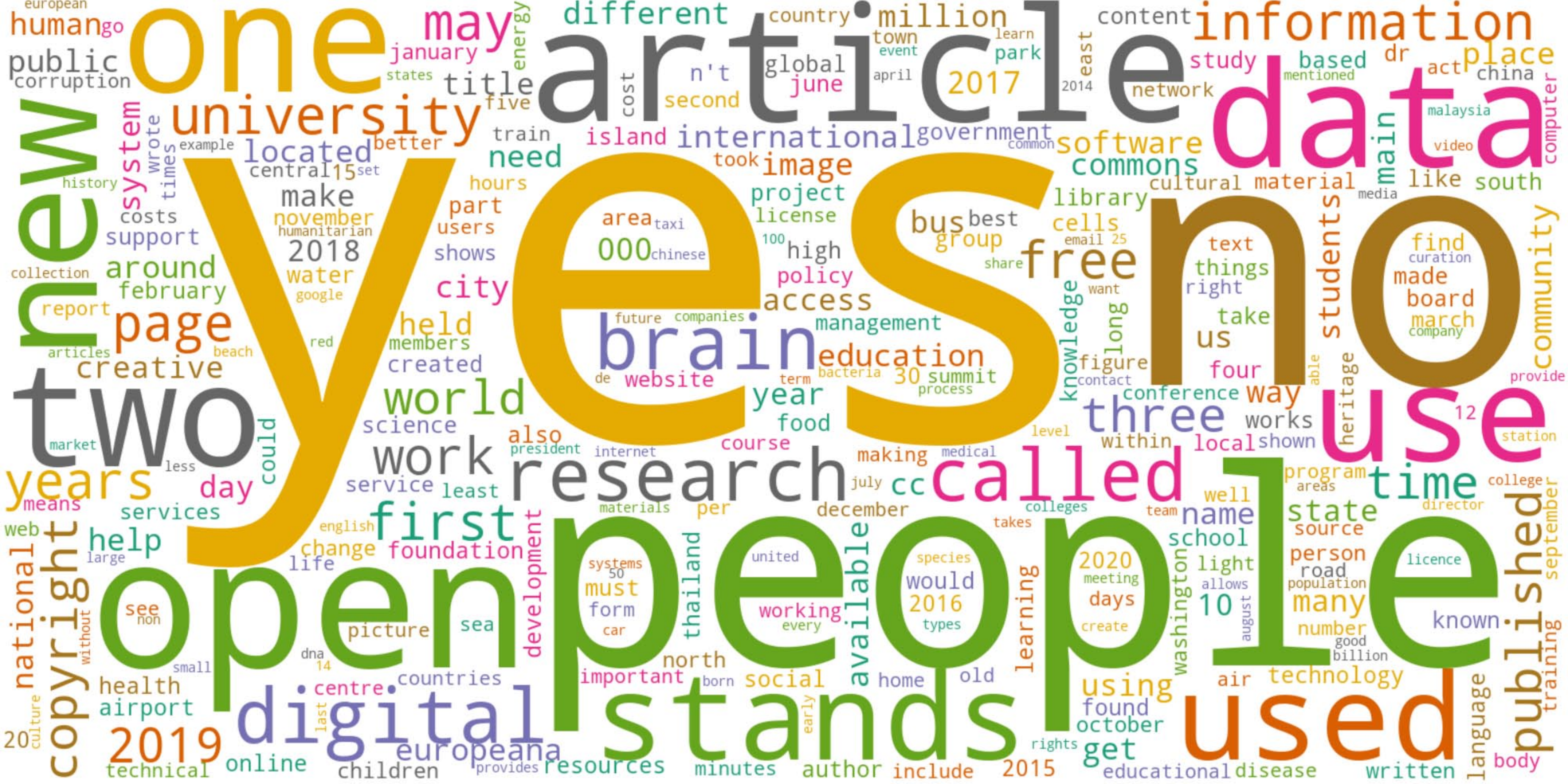}
    }
    \hspace{1em}
    \subfigure[Word cloud for document images.]{
    \includegraphics[width=.29\textwidth]{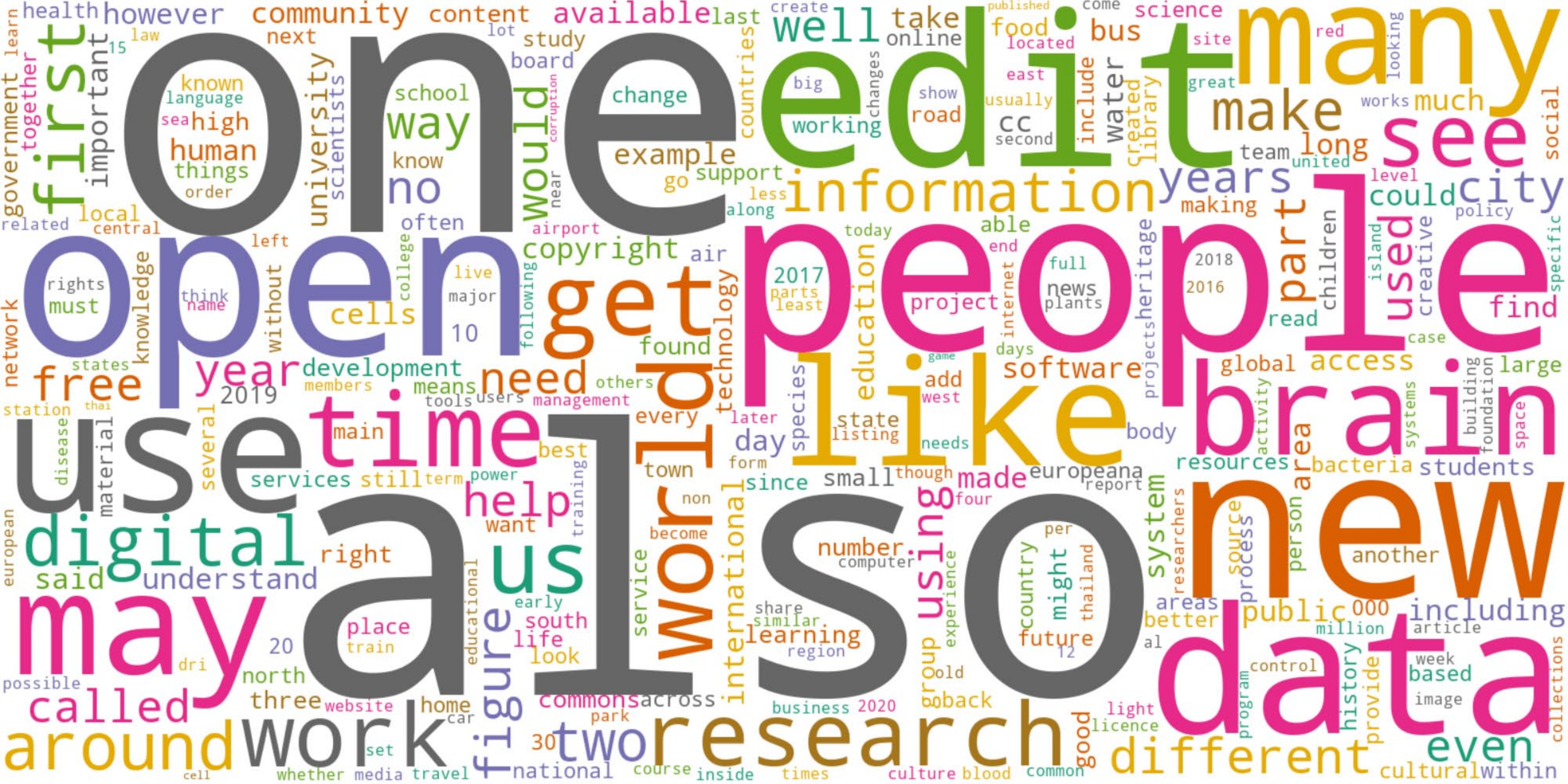}
    }
    \caption{Statistics of tokens in question, answer, document images of VisualMRC, TextVQA, and DocVQA datasets. Stop words were excluded from word clouds.
    }
    \label{fig:statistics}
\end{figure*}

\subsection{Statistics and Analysis}

We compared our dataset with the two representative VQA datasets with text in images: TextVQA and DocVQA.

\begin{table}[t!]
    \centering
        \scalebox{0.97}{
\footnotesize
    \tabcolsep=4pt 
    \begin{tabular}{l|rrr} \hline
          & TextVQA & DocVQA & VisualMRC \\ \hline
        Image type & \begin{tabular}{r} daily \\ scenes\end{tabular}& \begin{tabular}{r}industry \\ documents\end{tabular} & 
        \begin{tabular}{r}webpages\end{tabular} \\ \hline
        Num. images & 28,472 & 12,767 & 10,197 \\
        Num. questions & 45,536 & 50,000 & 30,562 \\
        Uniq. num. questions & 36,593 & 36,170 & 29,419 \\ \hline
        Perc. uniq. answers & 51.74 & 64.29 & 91.82 \\
        \hline
        Avg. len. questions & 8.12 & 9.49 & 10.55 \\
        Avg. len. documents & 12.17 & 182.75 & 151.46 \\
        Avg. len. answers & 1.51 & 2.43 & 9.53 \\ 
        \hline
    \end{tabular}
    }
    \caption{Statistics of datasets. 
    The percentages of unique answers and 
    average lengths of answers in TextVQA and DocVQA were calculated with the majority answers in the train and dev 
    sets.  The lengths of the questions and answers were measured by tokenizing them with NLTK. The lengths of the documents were counted in terms of OCR words.}
    \label{tab:statistics}
\end{table}

\subsubsection{Questions.}
\begin{figure}[t!]
    \centering
    \includegraphics[width=.35\textwidth]{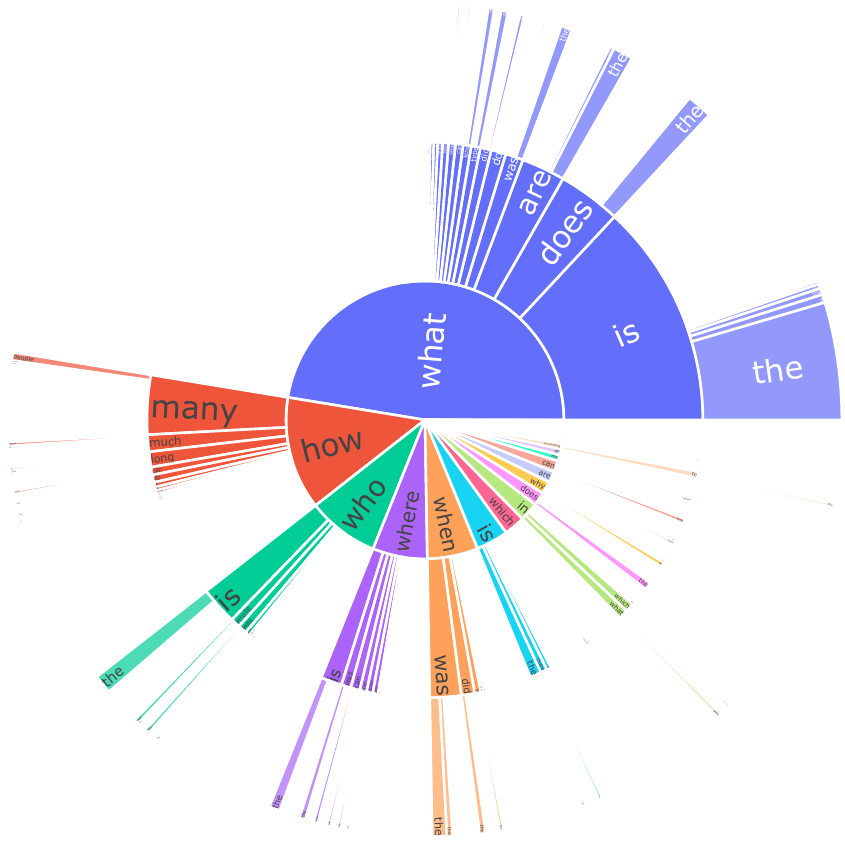}
    \caption{Distribution of questions by their first three words.}
    \label{fig:sunburst}
\end{figure}

Table~\ref{tab:statistics} shows that 
the percentage of unique questions in VisualMRC (96.3\%) is higher than in TextVQA (80.7\%) or DocVQA (72.3\%).
The average question length in VisualMRC is 10.55 tokens, larger than in TextVQA (8.12) or DocVQA (9.49).
Figure~\ref{fig:statistics}a shows that the distribution of question lengths is more long-tailed than in TextVQA and DocVQA. 
Figure~\ref{fig:statistics}d shows a word cloud that presents the question space is diverse. 
Figure~\ref{fig:sunburst} shows the first three words of the questions.
Compared with TextVQA and DocVQA, we can see that a variety of questions are included in the VisualMRC dataset.
Questions often start with ``what'' (42.0\%) and ``what is the'' (9.5\%), while their percentages are significantly lower than those in TextVQA (78.2\% and 22.9\%) and DocVQA (68.1\% and 58.2\%). 
Yes/no questions (starting with "is" (4.7\%), "are" (2.0\%), "can" (1.8\%), "do" (1.1\%), etc.) are also included in the dataset.

\subsubsection{Answers.}


Table~\ref{tab:statistics} shows that the percentage of questions having unique answers in VisualMRC (91.82\%) is significantly higher than those of TextVQA (51.74\%) and DocVQA (64.29\%).
The average answer length in VisualMRC is 9.53 tokens, 
significantly larger than in TextVQA (1.51) and DocVQA (2.43). 
Also, 
answers begin with ``yes'' in 10.04\%  and ``no'' in 2.67\% of the whole answers. These percentages are higher than those in TextVQA (4.90\% and 0.97\%) and DocVQA (0.12\% and  0.15\%).

\subsubsection{Document images.}

The average number of OCR words in an image of VisualMRC (151.46) and DocVQA (182.75) is significantly larger than TextVQA (12.17).
We also analyzed topics of 
the documents by using Latent Dirichlet Allocation (LDA)~\cite{BleiNJ03}, and found that VisualMRC covers a broad range of topics, science, travel, software, health, education, news, government, etc.~(see Table~\ref{tab:topics}),
while most of the documents in DocVQA relate to food and nutrition~\cite{Mathew_2021_WACV}.

\begin{table}[t!]
\scalebox{0.95}{
\tabcolsep=2pt
\footnotesize
\begin{tabular}{c|p{25em}}
\hline
\# & topic words \\ \hline
1 & brain cells figure people different like called body see children \\
2 & city understand many area south north world island east park \\ 
3 & get bus around road airport city station add listing train take \\ 
4 & use software copyright free information content work may \\ 
5 & water figure species bacteria plants food called different fish \\
6 & first university music film years london wikipedia history new \\
7 & like new people world technology make even time future way \\
8 & health humanitarian disease medical medicine cancer research \\
9 & open data digital research community education cc europeana \\
10 & government people said thailand corruption countries country \\
\hline
\end{tabular}
}
\caption{Ten topics inferred from 
document images by LDA. We treated the OCR words in an image as a document.}
\label{tab:topics}
\end{table}

\begin{figure}[t!]
    \centering
    \includegraphics[width=.35\textwidth]{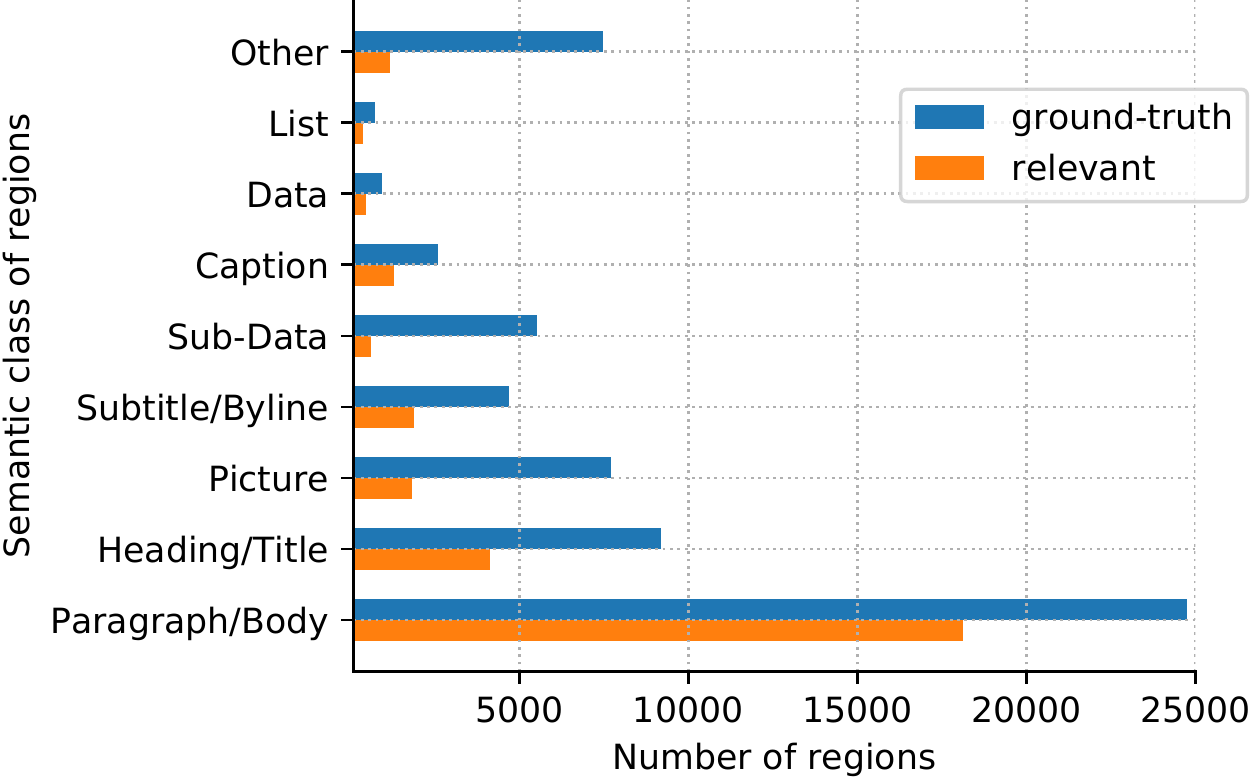}
    \caption{Total number of ground-truth and relevant ROIs in document images broken down by semantic class. 
    }
    \label{fig:regions}
\end{figure}

Moreover, unlike DocVQA and TextVQA, VisualMRC provides ROI annotations in images.  
Figure~\ref{fig:regions} shows the number of ROIs broken down into the nine semantic classes. The paragraphs and titles tend to be related to the question.
Also, 44.8\% of the document images contain picture regions and/or data regions such as tables and charts.

\begin{figure*}
    \centering
    \includegraphics[width=0.92\textwidth]{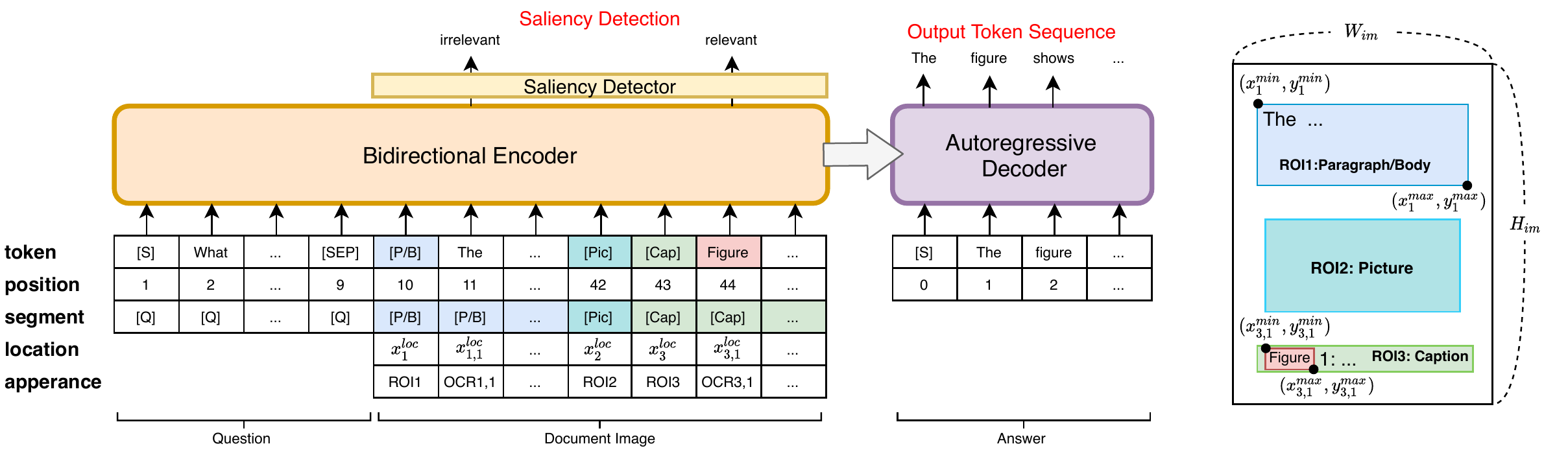}
    \caption{\textbf{Left}: Our encoder-decoder model architecture.
    A saliency detector that finds important tokens relevant to the question is trained at the same time the sequence-to-sequence
    task is being learned.
    A sequence of five embeddings is passed to the encoder.
    Special tokens such as \texttt{[P/B]} corresponding to the semantic classes of ROIs are used for the token and segment embeddings.
    \textbf{Right}: Example of ROIs and OCR tokens (e.g., $w_{3,1}=$ ``Figure'' in ROI3) in the document image. 
    Their relative locations are used in the location embeddings, and their visual features are considered in the appearance embeddings.}
    \label{fig:proposed_model}
\end{figure*}

\section{Proposed Model}

Our model consists of sub-modules for the ROI detection and OCR (\textsc{Subtasks} 1 and 2) and a main module for visual machine reading comprehension. In this section, we first explain the main module and then the sub-modules.

Our main module has a Transformer~\cite{VaswaniSPUJGKP17} architecture (see Figure~\ref{fig:proposed_model}).
Following the success of recent pre-trained encoder-decoder models such as BART~\cite{LewisLGGMLSZ20} and T5~\cite{RaffelSRLNMZLL20} in NLG tasks, we extended the models by 
learning the visual layout and content of the documents in a way that does not cause the 
forgetting of the pre-trained NLG ability.

The main module maps an input sequence to a sequence of embeddings, which is passed to the encoder. We extended the formalization of the input sequence and embeddings.
We also conducted saliency detection to find the tokens relevant to the question.

\subsection{Input Sequence}
The input sequence is formed from the tokenization results of the concatenation of a question and OCR words in ROIs, which are the outputs of the \textsc{Subtasks} 1 and 2. To consider the semantic roles of ROIs, we insert a special token \texttt{[L$_{r_i}$]} corresponding to the semantic class label $l_i$ of the $i$-th ROI $r_i$ (such as \texttt{[P/B]}; see Figure~\ref{fig:proposed_model}) just before the sequence of OCR tokens $\{w_{r_i,1}, \ldots, w_{r_i,M}\}$ in $r_i$:
\begin{equation}
\nonumber
x^{\rm token} = \biggl\{
\begin{aligned}
&{\rm \texttt{[S]}}, q_1, ..., q_m, {\rm \texttt{[SEP]}}, [{\rm \texttt{L}}_{r_1}], w_{1,1}, ..., w_{1,M}, \\
&\hspace{.5em}   
[{\rm \texttt{L}}_{r_2}], \dots,
[{\rm \texttt{L}}_{r_N}], 
w_{r_N,1}, ..., w_{r_N,M}
\end{aligned}
\biggr\},
\end{equation}
where 
\texttt{[S]} is a 'question:' (\texttt{<s>}),
and \texttt{[SEP]} is a 'context:' (\texttt{</s>})
in the case 
we use T5 (BART) as the base model.

\subsection{Input Embeddings}

The input embeddings of the ROI and OCR tokens, which are passed to the encoder, consist of a segment $z^{{\rm seg}}$, a location within the image $z^{{\rm loc}}$, and an appearance $z^{{\rm app}}$ 
in addition to a token $z^{{\rm token}}$ and a position within the input sequence $z^{{\rm pos}}$. 
In total, the fused embedding $z_k \in \mathbb{R}^{H}$ at the $k$-th position in the sequence, $z_k$, 
is given as:
\begin{equation}
\nonumber
z_k = {\rm LN}(z^{{\rm token}}_k + z^{{\rm pos}}_k + z^{{\rm seg}}_k + z^{{\rm loc}}_k + z^{{\rm app}}_k)
\end{equation}
where LN($\cdot$) is a layer normalization~\cite{BaKH16}.
Note that the $z^{{\rm pos}}_k$ of T5 is set to a zero vector since T5 uses relative position embeddings in self-attention~\cite{shaw2018self} instead of the absolute position embeddings. 
Also note that the segment, location, and appearance embeddings are not used by the decoder, and those of the special tokens \texttt{[S]} and \texttt{[SEP]} and question tokens for the encoder are set to zero vectors.

We explain the three additional embeddings below.

\subsubsection{Segment embeddings.}
To convey the semantics of the document structure to the model more explicitly, we introduce a learnable segment embedding vector $z^{{\rm seg}}_k \in \mathbb{R}^{H}$ for each ROI class, indicating which ROI the $k$-th token in the input sequence comes from.

\subsubsection{Location embeddings.}
We introduce a location embedding $z^{{\rm loc}}_k \in \mathbb{R}^{H}$ that denotes the relative location of the $k$-th token (corresponding to a ROI or an OCR token) within the input image.
We use a 4-dimensional location feature based on the token’s relative bounding box coordinates:
\begin{equation}
\nonumber
x^{{\rm loc}}_k = [ x_k^{{\rm min}}/W_{{\rm im}}, y_k^{{\rm min}}/H_{{\rm im}}, x_k^{{\rm max}}/W_{{\rm im}}, y_k^{{\rm max}}/H_{{\rm im}}],
\end{equation}
where $(x_k^{{\rm min}}, y_k^{{\rm min}})$ and $(x_k^{{\rm max}}, y_k^{{\rm max}})$ are the coordinates of the top-left and bottom-right corners of the $k$-th token's bounding box, and $W_{{\rm im}}$ and $H_{{\rm im}}$ are the image width and height, respectively (see Figure~\ref{fig:proposed_model}). 
$x^{{\rm loc}}_k$ is passed to a 1-layer FFN to obtain the locations embedding $z^{{\rm loc}}_k$.

\subsubsection{Appearance embeddings.}
To consider the appearance of the ROIs and OCR tokens, 
we incorporate their visual features into the input embeddings.
The image corresponding to the bounding box of the $k$-th token 
is fed into a Faster R-CNN~\cite{ren2015faster}
to obtain 2048-dimensional fc7 appearance features, $z^{\rm fc7}_k \in \mathbb{R}^{2048}$.
Then, the ReLU activation of the feature $z^{\rm fc7}_k$ is passed to a 1-layer FNN to obtain 
the appearance embedding $z^{{\rm app}}_k$.

\subsection{Saliency Detection}
To find the tokens relevant to the question,
we calculate the relevance of each 
token with the outputs of the encoder,
\begin{equation}
\nonumber
P_{i,j}
= {\rm sigmoid}({w^{s}}^\top h_{w_{i,j}} + b^s),
\end{equation}
where 
$h_{w_{i,j}}$ is the encoder hidden state in the last layer corresponding to $w_{i,j}$ (the $j$-th OCR token in the $i$-th ROI).   
$w^{s} \in \mathbb{R}^H$, $b^{s} \in \mathbb{R}$ are learnable weights.

\subsubsection{Saliency loss.} We introduce a saliency loss to better supervise the determination of which tokens are required to answer the question. However, a reference label for each token is not given. To deal with this problem, we make pseudo reference labels $s$ by aligning the OCR tokens and answers. The reference label is set to 1 if the following two conditions are met: (i) an OCR token exists in the answer, and (ii) an OCR token belongs to the relevant 
ROIs; otherwise, it is set to 0. The saliency loss is defined as:
\begin{equation}
\nonumber
L_{\rm sal} = - \frac{1}{NM} \sum_i^{N} \sum_j^{M} \left(
\begin{array}{l} 
   s_{i, j} \log P_{i, j}
   +    \\
  (1 - s_{i, j}) \log (1 -P_{i, j})
\end{array} \right)
\end{equation}


\subsubsection{Multi-task learning.}
Our main module is trained by minimizing the weighted sum of the two losses:
\\
$L_{\rm multi} = L_{\rm nll} + \gamma_{\rm sal} L_{\rm sal}$, 
where $\gamma_{\rm sal}$ is a hyper-parameter to balance the losses, and $L_{\rm nll}$
is the negative log-likelihood loss in the sequence-to-sequence learning. 

\subsection{Sub-modules for ROI detection and OCR}
 
Using a different Faster R-CNN than the one for appearance embedding,
up to 100 detected objects with the highest score are selected 
for each document image. 
This sub-module is trained independently of the main module.  Also, a built-in OCR system such as Tesseract~\cite{Smith07} 
without any fine-tuning is used to extract OCR words 
from each 
ROI.

\section{Experiments}

\begin{table*}[t!]
    \centering
        \scalebox{0.95}{
\footnotesize
    \tabcolsep=3pt
    \begin{tabular}{l|ccc|c|c|c|c|c|c|c|c} \hline
        Model & OCR & Q & V & BLEU-1 & BLEU-2 & BLEU-3 & BLEU-4 & METEOR & ROUGE-L & CIDEr & BERTscore \\ \hline 
        M4C-Q & & \checkmark & & 20.2 & 13.0 & 8.9 & 6.1 & 9.8 & 20.9 & 58.3 & 85.1\\ 
        M4C-Visual & &  \checkmark & \checkmark & 20.7 & 13.3 & 9.2 & 6.3 & 10.1 & 21.8 & 61.0 & 85.3 \\
        M4C-Text & \checkmark & \checkmark & & 26.7 & 17.4 & 11.8 & 8.8 & 11.6 & 26.9 & 88.3 & 85.9 \\
        M4C & \checkmark & \checkmark & \checkmark & 29.2 & 20.1 & 14.4 & 10.3 & 12.8 & 28.1 & 98.6 & 86.1 \\ 
         \hline 
        T5-Q & & \checkmark & & 31.2 & 25.9 & 22.6 & 20.0 & 18.5 & 29.6 & 155.0 & 87.5 \\  
        T5-Text & \checkmark & \checkmark & & 53.0 & 48.2 & 44.5 & 41.5 & 31.7 & 53.0 & 318.6 & 90.5 \\  
        BART-Q & & \checkmark & & 31.8 & 25.7 & 21.9 & 19.0 & 15.0 & 27.7 & 140.5 & 73.0   \\
        BART-Text & \checkmark & \checkmark & & 50.6 & 44.4 & 39.9 & 36.4 & 28.8 & 48.7 & 278.3 & 90.1 \\ \hline       
        LayoutT5  & \checkmark & \checkmark & \checkmark & \textbf{56.0} & \textbf{50.8} & \textbf{46.7} & \textbf{43.4} & 34.6 & \textbf{54.6} & \textbf{335.9} & \textbf{90.8} \\
        LayoutT5 w/o Saliency Detection & \checkmark & \checkmark & \checkmark & 55.8 & 50.7 & 46.6 & 43.3 & \textbf{34.9} & 54.4 & 335.1 & 90.7 \\  
        LayoutBART  & \checkmark & \checkmark & \checkmark & \textbf{53.0} & \textbf{46.8} & \textbf{42.3} & \textbf{38.7} & \textbf{31.9} & \textbf{52.8} & \textbf{309.9} & \textbf{90.7}  \\ 
        LayoutBART w/o Saliency Detection & \checkmark & \checkmark & \checkmark & 52.0 & 45.8 & 41.3 & 37.7 & 31.3 & 52.8 & 302.8 & 90.6 \\  \hline \hline 
        LayoutT5$_{\rm LARGE}$ & \checkmark & \checkmark & \checkmark & 57.2 & 52.1 & 48.1 & 44.9 & 37.3 & 57.1 & 364.2 & 91.3 \\
        LayoutBART$_{\rm LARGE}$ & \checkmark & \checkmark & \checkmark & 57.2 & 51.2 & 46.7 & 43.0 & 36.1 & 57.0 & 346.0 & 91.5 \\ \hline 
    \end{tabular}
    }
    \caption{Main evaluation results for different methods that leverage OCR, Question (Q) and Visual (V). 
    }
    \label{tab:automatic_evaluation}
\end{table*}

We conducted the evaluation experiments with our VisualMRC dataset. 
We did not use DocVQA because it does not provide ROI annotations in images and does not focus on generating answers. We used BART~\cite{LewisLGGMLSZ20} and T5~\cite{RaffelSRLNMZLL20} as our initial models of our proposed model. We fine-tuned them with the dataset, calling them \textbf{LayoutBART} and \textbf{LayoutT5}, respectively.  

\subsection{Experimental Setup}

\subsubsection{Evaluation settings.}

In the \textbf{end-to-end} evaluation setting (corresponding to \textsc{Task} 1), we conducted ROI detection with a Faster R-CNN (\textsc{Subtask} 1) and used Tesseract~\cite{Smith07} to extract OCR words from each ROI (\textsc{Subtask} 2); while in the \textbf{main} evaluation setting, we used the ground-truth ROIs (manually annotated bounding boxes and semantic class labels) and the OCR words for the ROIs provided in the VisualMRC dataset.  Our model and the baselines were trained on the ground-truth ROIs
for both the evaluation settings.
We evaluated our model in the main setting unless otherwise stated. 
Note that the performance in the end-to-end setting is important for application scenarios in the real-world.

When making comparisons with \textbf{human performance} under the main setting, we first randomly picked 3,000 QA pairs (on 1,000 images) from the test set. Next, one reliable worker answered the questions about the images. Then, another reliable worker validated the correctness of the answers created by the first reliable worker. Finally, 
we compared the original answers and the answers created and validated by the reliable workers.

\subsubsection{Baselines.}
We used a state-of-the-art model for text-based VQA, \textbf{M4C} \cite{HuSDR20}, that takes OCR tokens, question, and ROI features as input and generates an answer word-by-word sequentially. 
Note that the whole architecture of M4C except the input embeddings (OCR, question, and ROI) were not pre-trained.
In addition, we used the fine-tuned T5 and BART without visual information (\textbf{T5-Text} and \textbf{BART-Text}). They correspond to our models without the segment, location, and appearance embeddings and without saliency detection. We used T5$_{\rm BASE}$ and BART$_{\rm BASE}$ unless otherwise stated.
We also used the ones that take a question only as input (\textbf{-Q}) and that 
take visual features only as input (\textbf{-Visual}).

\subsubsection{Evaluation metrics.}
Following the previous generative MRC and image captioning tasks, we used BLEU \cite{papineni2002bleu}, METEOR \cite{denkowski2014meteor}, ROUGE-L \cite{lin2004rouge}, and CIDEr \cite{vedantam2015cider} to assess the quality of the generated answers. These scores were calculated with the coco-caption toolkit.  We also used the F1 score of BERTScore, which is highly correlated with human judgment~\cite{ZhangKWWA20}.


\subsection{Implementation Details}
 We implemented all the models in PyTorch 
 and experimented on eight NVIDIA Quadro RTX 8000 GPUs.

\subsubsection{Main module.} 
We implemented our main module based on the BART~\cite{LewisLGGMLSZ20} and T5~\cite{RaffelSRLNMZLL20} of huggingface Transformers~\cite{Wolf2019HuggingFacesTS}.
We mainly used BART$\rm_{BASE}$ (six layers with 768 hidden size) and T5$_{\rm BASE}$ (12 layers with 768 hidden size) to initialize our models.
The following settings for the BASE version are the same as those for the LARGE version of BART (12 layers with a hidden size of 1024) and T5 (24 layers with a hidden size of 1024).

The balancing parameter $\lambda_{\rm sal}$ was set to $1.0$.
During training, we used a batch size of 32, and trained for a maximum of seven epochs. Our model was trained using the Adam optimizer \cite{KingmaB15} with a learning rate of 3e-5. The best model in terms of ROUGE-L was selected using the validation set. 
When an OCR word is tokenized into sub-word tokens, the bounding box coordinates of a sub-word token are the same as those of its whole word as in LayoutLM~\cite{XuLCHWZ20}.

For the appearance embeddings, we used a Faster R-CNN~\cite{ren2015faster} with a ResNet-101~\cite{HeZRS16} backbone pre-trained on Visual Genome~\cite{KrishnaZGJHKCKL17}, where we used the code and model of M4C\footnote{\url{https://github.com/ronghanghu/pythia}}~\cite{HuSDR20}.
Then, the fc7 weights of the Faster R-CNN were only fine-tuned during the training of the main module with the VisualMRC dataset.
For the saliency detection, we used a label smoothing technique \cite{szegedy2016rethinking} to smooth positive labels to 0.9.

 \subsubsection{Sub-module for ROI detection.} 
We trained another Faster-RCNN with a ResNet-101 backbone with VisualMRC independently of the main module for three epochs, with a batch size of 16 and the Adam optimizer. The starting learning rate was 1e-3. Standard anchor scales of [8, 16, 32] and anchor ratios of [0.5, 1.0, 2.0] were used.  

\subsubsection{M4C.}
We implemented M4C and its variant based on 
the above-mentioned authors' code.
To enrich the OCR token representation of M4C, we replaced FastText~\cite{bojanowski2017enriching} with BERT$_{\rm BASE}$ of huggingface Transformers.
We used the ROIs of the VisualMRC dataset as the detected objects to be handled in M4C.

\subsection{Evaluation Results}

\subsubsection{Do our models outperform other models?}
Table \ref{tab:automatic_evaluation} shows that our models outperformed the baselines on all metrics. This indicates that the additional learning of the visual layout and content of documents improves performance. 
M4C, a non pre-trained VQA model, performed significantly worse than BART and T5.
This indicates that the transfer of their pre-trained NLG ability to the VisualMRC task is useful in generating answers.  
\citet{Mathew_2021_WACV} also reported that
BERT~\cite{DevlinCLT19} outperformed M4C in the DocVQA task.
We also found significant performance improvements on all metrics when the model parameters were increased from BASE to LARGE. 
Moreover, we noticed that the models that disregarded the visual information and OCR tokens performed worse than the full models.  

\subsubsection{Does multi-task learning with saliency detection improve the performance?}
Table \ref{tab:automatic_evaluation} shows that our models (jointly trained with the saliency detector) outperformed the other models that did not have the saliency detector, except METEOR of LayoutT5. 
The improvements in LayoutT5 were smaller than those in LayoutBART because T5's pre-training includes an MRC task~\cite{RajpurkarZLL16} that is similar to saliency detection in terms of extracting of important pieces from a document.


\begin{table}[t!]
    \centering
        \scalebox{0.9}{
\footnotesize
    \tabcolsep=3pt
    \begin{tabular}{l|c|c|c|c|c} \hline
        Model & BLEU-4 & METEOR & ROUGE-L & CIDEr & BERTscore \\ \hline 
        M4C & 10.2 & 12.7 & 28.0 & 97.6 & 86.1 \\
        T5-Text & 38.6 & 29.8 & 50.2 & 297.6 & 90.0 \\
        \hspace{0.15cm} w/o ROI det & 37.5 & 28.8 & 48.6 & 284.3 & 89.5 \\
        BART-Text & 34.6 & 27.5 & 47.3 & 265.6 & 90.0 \\ 
        \hspace{0.15cm} w/o ROI det & 33.2 & 27.2 & 46.7 & 258.6 & 89.7 \\
        \hline 
        LayoutT5 & \textbf{41.0} & \textbf{33.2} & \textbf{52.2} & \textbf{317.9} & \textbf{90.3}  \\  \hspace{0.15cm} 
        w/o ROI det & 39.1 & 31.0 & 49.3 & 292.8 & 89.8  \\
        LayoutBART & \textbf{36.4} & \textbf{30.5} & \textbf{50.5} & \textbf{293.9} & \textbf{90.4} \\ 
        \hspace{0.15cm} w/o ROI det & 33.8 & 29.6 & 48.6 & 277.3 & 90.0 \\ 
        \hline 
    \end{tabular}
    }
    \caption{
    Performance in the end-to-end setting.}
    \label{tab:end-to-end}
\end{table}

\subsubsection{Do our models also perform well in the end-to-end setting?}
Table \ref{tab:end-to-end} shows that our models also outperformed the baselines on all metrics in the end-to-end setting that is important for application scenarios in the real-world.
The performances in the end-to-end setting was not significantly decreased compared with those in the main setting.
But there is still room for improving ROI detection (the mean Average Precision was 7.86\%). 
This was comparable to the performance of a Faster R-CNN (5.1\%) reported by \citet{SotoY19} on a document layout analysis dataset~\cite{tkaczyk2014grotoap2}. 

Furthermore, we compared the models that directly read the images (w/o ROI detection). Table \ref{tab:end-to-end} shows 
that ROI detection was effective.
This is because our OCR system fails to read tokens in the correct order when reading complicated (multi-column) documents, 
and ROI detection enables our model to utilize visual layout information.




\begin{table}[t!]
    \centering
    \scalebox{0.9}{
\footnotesize
    \tabcolsep=3pt
    \begin{tabular}{l|c|c|c|c|c} \hline
        Model & BLEU-4 & METEOR & ROUGE-L & CIDEr & BERTscore \\ \hline 
        T5-Text & 41.5 & 31.7 & 53.0 & 318.6 & 90.5\\  
        + lbl & 42.9 & 32.5 & 53.2 & 321.0 & 90.5 \\
\hspace{0.15cm} + seg & 43.6 & 32.8 & 53.3 & 320.7 & 90.5 \\
\hspace{0.3cm} + loc & \textbf{44.1} & 33.5 & 53.7 & 325.2 & 90.5  \\
\hspace{0.45cm}  + app & 43.3 & \textbf{34.9} & \textbf{54.4} & \textbf{335.1} & \textbf{90.7} \\ \hline 
        BART-Text & 36.4 & 28.8 & 48.7 & 278.3 & 90.1 \\  
       + lbl & 37.6 & 30.3 & 50.7 & 293.7 & 90.3 \\
\hspace{0.15cm} + seg & 37.8 & 30.3 & 50.9 & 296.0 & 90.4
\\
\hspace{0.3cm}  + loc & \textbf{38.1} & 30.3 & 51.4 & 296.3 & 90.5 \\
\hspace{0.45cm} + app & 37.7 & \textbf{31.3} & \textbf{52.8} & \textbf{302.8} & \textbf{90.6} \\ \hline 
    \end{tabular}
    }
    \caption{Performance in the case of
    inserting ROI class labels (lbl) and adding other embeddings (seg, loc, and app).
    }
    \label{tab:modify_input}
\end{table}

\subsubsection{Is modifying the input sequence and embeddings effective?}
Table \ref{tab:modify_input} shows the results of modifying the input sequence and embeddings of the baselines (BART-Text and T5-Text). First, inserting the region class labels (lbl) before the OCR token sequence consistently improved almost all the metrics except BERTscore of T5-Text. Second, adding the segment (seg) and location (loc) embeddings also improved performance. Third, using the appearance embedding (app) improved the performance except in terms of BLEU-4; this observation is in line with previous studies \cite{le-hoi-2020-video,li2020bridging}.

\begin{table}[t!]
    \centering
    \scalebox{0.82}{
\footnotesize
    \tabcolsep=3pt
    \begin{tabular}{c|c|c|c|c|c} \hline
         & BLEU-4 & METEOR & ROUGE-L & CIDEr & BERTscore \\ \hline 
        Heading/Title & 37.9/\textbf{42.8} & 29.8/\textbf{32.5} & 49.9/\textbf{52.6} & 289.2/\textbf{315.4} & 89.9/\textbf{90.3} \\ 
        Paragraph/Body & 42.7/\textbf{44.1} & 32.3/\textbf{35.0} & 54.0/\textbf{55.1} & 328.1/\textbf{340.9} & 90.6/\textbf{90.8} \\
        Subtitle/Byline & 39.6/\textbf{46.3} & 29.9/\textbf{33.8} & 48.0/\textbf{52.6} & 314.2/\textbf{353.1} & 90.0/\textbf{90.8} \\
        Picture & 25.9/\textbf{32.0} & 24.8/\textbf{29.8} & 44.9/\textbf{49.0} & 242.6/\textbf{282.7} & 89.4/\textbf{90.3} \\
        Caption & 31.2/\textbf{41.1} & 28.0/\textbf{33.1} & 50.3/\textbf{55.5} & 289.4/\textbf{344.0} & 89.6/\textbf{91.0} \\
        List & 35.7/\textbf{39.0} & 30.4/\textbf{33.1} & 48.1/\textbf{50.4} & 282.5/\textbf{307.0} & 90.0/\textbf{90.7} \\
        Data & 31.8/\textbf{32.7} & 26.1/\textbf{29.3} & 42.2/\textbf{46.4} & 248.5/\textbf{287.0} & 88.9/\textbf{89.6} \\
        Sub-Data & 30.1/\textbf{41.4} & 26.4/\textbf{32.4} & 42.8/\textbf{50.6} & 236.3/\textbf{315.8} & 88.9/\textbf{90.6} \\
        Other &  34.1/\textbf{41.5} & 28.1/\textbf{32.5} & 48.4/\textbf{51.7} & 260.6/\textbf{290.0} & 89.8/\textbf{90.5} \\ 
        \hline
    \end{tabular}
    }
    \caption{Performance of T5/LayoutT5 
    broken down by semantic class.}
    \label{tab:performence_type}
\end{table}

\subsubsection{On which classes of ROI does our model work well?}
Table \ref{tab:performence_type} shows the performance broken down by semantic class according to whether it is included in the relevant ROIs.
LayoutT5 performed better than T5-Text on all metrics and all semantic classes. Particularly, LayoutT5 showed significantly improvements on the Picture, Caption, Sub-Data, and Other classes. This indicates that our model was especially effective at understanding vision-related data. However, both T5-Text and LayoutT5 underperformed on the Picture and Data classes compared with the other classes; determining ways of improving performance on these classes will be our future work.

\begin{table}[t!]
    \centering
        \scalebox{0.9}{
\footnotesize
    \tabcolsep=3pt
    \begin{tabular}{c|c|c|c|c|c} \hline
         & BLEU-4 & METEOR & ROUGE-L & CIDEr & BERTScore \\ \hline 
        LayoutT5 & \textbf{42.1} & 35.6 & 54.5 & 344.1 & 90.9 \\ 
        LayoutBART & 40.6 & 34.6 & 55.2 & 329.1 & 91.2   \\ \hline         
        Human & 39.6 & \textbf{41.0} & \textbf{57.9} & \textbf{370.3} & \textbf{91.9}  \\ \hline 
    \end{tabular} 
    }
    \caption{Human performance compared with those of our models in the end-to-end setting on the sampled test set. The architectures of LayoutT5 and LayoutBART were the LARGE versions.}
    \label{tab:human_performance}
\end{table}

\begin{table}[t!]
    \centering
    \footnotesize
    \begin{tabular}{c|c} \hline 
        Model & Avg. 
        Time \\ \hline 
        T5-Text & 0.1812 \\
        LayoutT5 & 0.2253 \\ 
        LayoutT5$_{\rm LARGE}$ & 0.4489 \\ \hline 
    \end{tabular}
    \caption{Average time (sec.) to answer a single question.}
    \label{tab:average_time}
\end{table}

\subsubsection{Do our models outperform than humans?}
Table~\ref{tab:human_performance} compares the performance of our best models (LayoutT5$_{\rm LARGE}$ and LayoutBART$_{\rm LARGE}$) with human performance in the end-to-end setting.
Our models achieved the highest BLEU-4, but the human level of performance was significantly higher on the other metrics. This indicates that there is still a performance gap between humans and the best machine. 

\subsubsection{How fast is our model?}

Table~\ref{tab:average_time} shows the average time to answer a single question from our models and the baselines with a NVIDIA Quadro RTX 8000 GPU. LayoutT5 needs to obtain a number of appearance embeddings for OCR tokens; however, LayoutT5 did not slow down significantly compared with T5-Text. The Faster R-CNN for appearance embeddings ran fast because it does not need to propose ROIs.  LayoutT5$_{\rm LARGE}$ ran much slower because it has about 3.5 times the parameters of LayoutT5.

\subsection{Output Example} 

\begin{figure}[t!]
    \includegraphics[width=.48\textwidth]{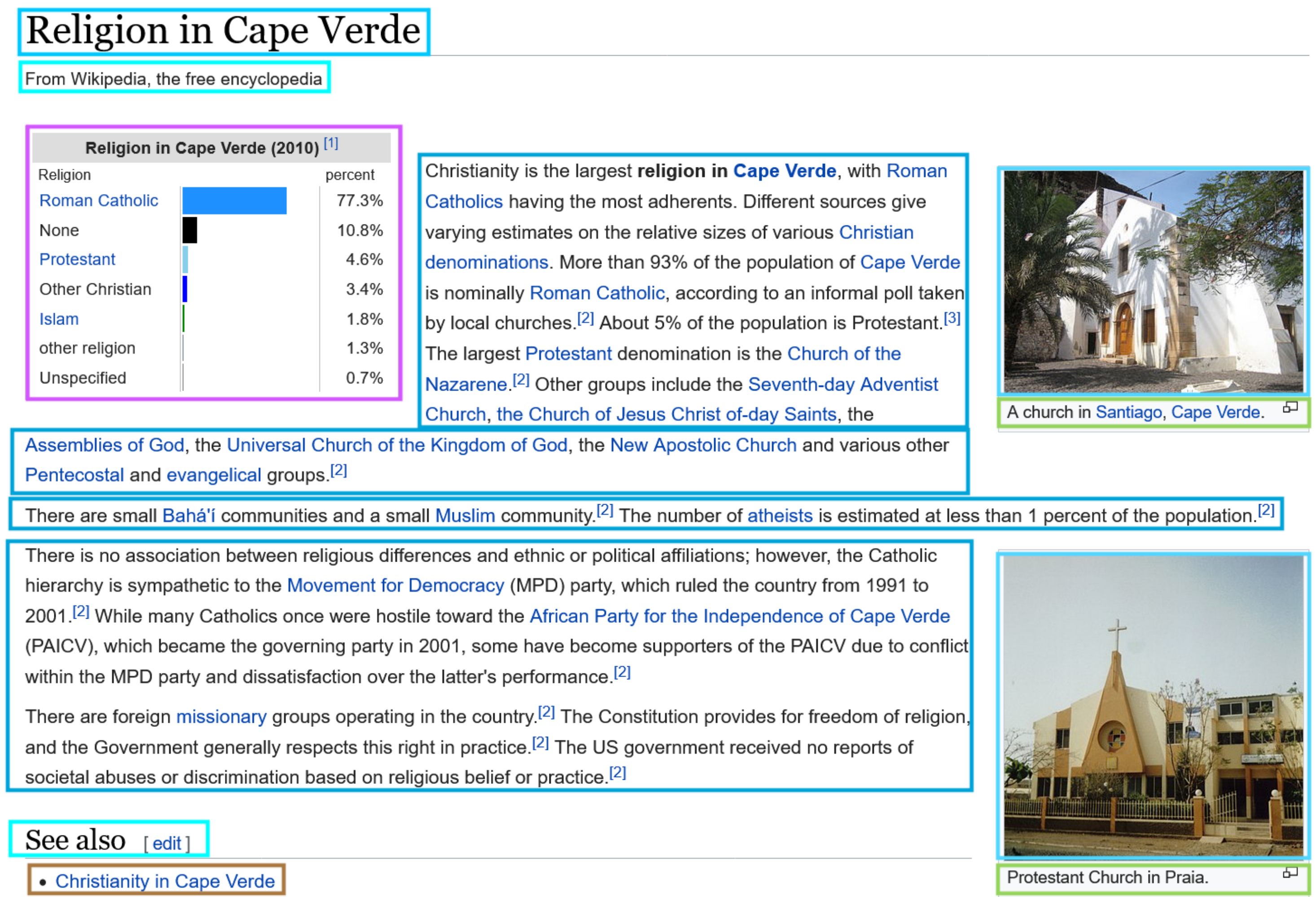}
    \footnotesize
    \tabcolsep=0pt
    \scalebox{0.95}{
    \begin{tabular}{p{28em}}
        \\
{\bf Q}: what is the percentage of roman catholics in cape verde? \\  
{\bf GT}: the percentage of roman catholics in cape verde is 77.3\%. \\
{\bf M4C}: the percentage of young women in cape town are about 54\% of western somalia \\
{\bf T5-Text}: percentage of roman catholics in cape verde is less than 1 percent. \\   
{\bf LayoutT5}: the percentage of roman catholics in cape verde is 77.3\%. \\
    \end{tabular}
    }
    \caption{Correct example generated by LayoutT5. GT denotes the ground-truth answer. The image was sourced from \url{https://en.wikipedia.org/wiki/Religion_in_Cape_Verde}.}
    \label{fig:sample_3}
\end{figure}

Figure \ref{fig:sample_3} shows an example of answers generated by the baselines and our model. 
In response to the question about the percentage of the Roman Catholics in Cape Verde, our model was able to understand that the first row of the visual data table contains the information required to answer the question (``77.3\%'') and generate the same answer as the ground-truth. T5-Text, which 
does not take into account the visual layout of the document images, was distracted by another percentage representation (``less than 1 percent'').

\begin{figure}[t!]
\centering
    \footnotesize
    \includegraphics[width=.38\textwidth
    ]{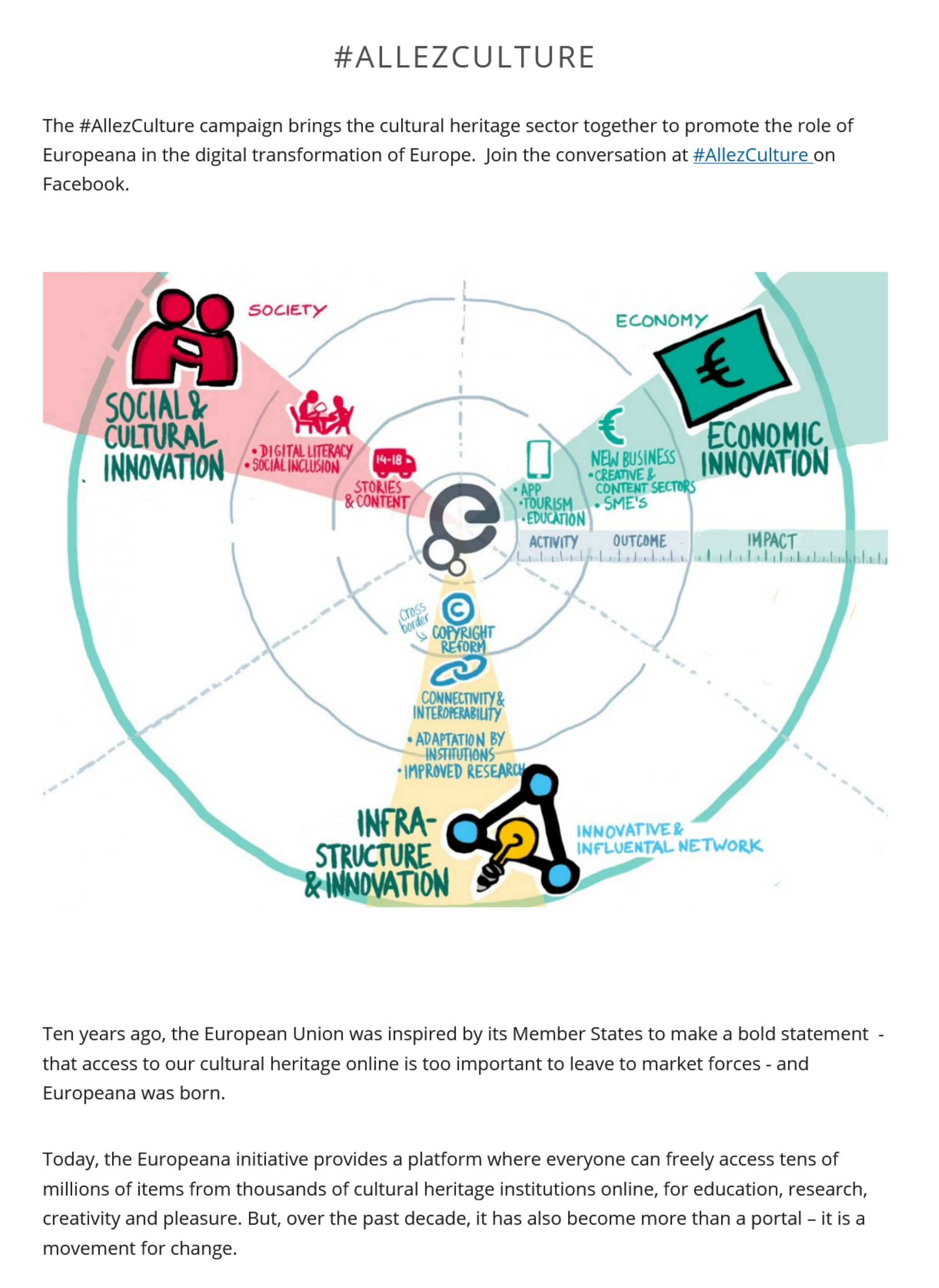}\\
    \tabcolsep=3pt
    \scalebox{0.95}{
    \begin{tabular}{p{28em}}
        {\bf Q}: in the graphic, what is the colour of economy? \\
        {\bf GT}: green \\ 
        {\bf M4C}: the colour of graphics is red. \\       
        {\bf T5-Text}: ay \\       
        {\bf LayoutT5}: economy is yellow. \\ 
        \\
    \end{tabular}
    }
\caption{Incorrect example. 
The image was sourced from \url{https://pro.europeana.eu/page/allezculture}.}
\label{fig:sample_notcorrect}
\end{figure}

Figure~\ref{fig:sample_notcorrect} shows an incorrect example, 
where the question is about the color of an object in the diagram. The proposed and baseline models could not identify the object related to the question and correctly answer its color. This indicates that further research should be conducted on detecting objects in diagrams within the document image.

\section{Related Work and Discussion}

\subsubsection{Transfer learning for vision and language.}

Recent Transformer-based vision and language models, pre-trained with large-scale image captioning and VQA datasets, have achieved state-of-the-art performances on vision and language tasks
~\cite{ChenLYKAGCL20,LiDFGJ20,LuGRPL20,ZhouPZHCG20}.
Moreover, as a pre-trained model for document layout analysis, \citet{XuLCHWZ20} proposed LayoutLM, which models interactions between text and layout information across grayscale scanned document images. It takes OCR words and their appearance as inputs and
performs well in form and receipt understanding and in document image classification.
But it cannot consider the appearance of visual content such as charts that can be handled with M4C and our models.

The NLG ability is important for the VisualMRC task, but the above pre-trained vision and language models, including LayoutLM, are not pre-trained on text-to-text tasks.
Sequence-to-sequence models pre-trained with large-scale text corpora have been successful in NLG tasks, 
so we decided to use BART and T5 as the base models and modified them into vision and language models.

\subsubsection{Document layout analysis.}
Neural networks have been recently used for page segmentation in order to split a document image into ROIs and to recognize the role of each ROI. \citet{YangYAKKG17} treated the task as a pixel-by-pixel classification problem. \citet{KattiRGBBHF18} treated each document page as a two-dimensional grid of characters and predicted a segmentation mask and bounding boxes. \citet{SotoY19} proposed an adaptation of the Faster R-CNN object detection model~\cite{ren2015faster}, with the addition of contextual features such as page numbers and ROI positions and sizes. 
We also used a Faster R-CNN to detect the bounding boxes and semantic classes of ROIs. To further improve the accuracy in the end-to-end VisualMRC setting, it will be important to improve the document layout analysis.

\section{Conclusion}

This study posed visual machine reading comprehension as a novel vision and language task.  Compared with existing VQA datasets such as TextVQA and DocVQA, our VisualMRC dataset focuses more on developing NLU and NLG abilities on various document images.
Our dataset contains 30,000+ questions defined on 10,000+ images of contemporary born-digital webpages on multiple domains. It requires a system to be able to read and reason about multiple pieces of text and non-text data in images and to generate abstractive answers.
We believe that this dataset will contribute to the development of intelligent assistant agents that can read real-world documents. 

Our visual machine reading comprehension models are based on encoder-decoder models pre-trained on large-scale text corpora, such as BART and T5, and they additionally learn the visual layout and content of document images.
Our models outperformed BART and T5 simply fine-tuned with only textual information and M4C, a state-of-the-art model for text-based VQA datasets that takes question, OCR tokens, and visual features of documents.
The key to its success is transferring the pre-trained NLG capability to the visual machine reading comprehension task by adding embeddings and an auxiliary saliency detection task for learning visual information in a way that does not cause catastrophic forgetting. Moreover, our approach can be easily applied to other pre-trained encoder-decoder models.
Our future work will involve exploring more effective pre-training methods for this task and improving the understanding of data objects such as tables, charts, and diagrams.

\bibliography{references}

\end{document}